\documentclass[11pt]{article}
\usepackage[]{acl}
\usepackage{times}
\usepackage{latexsym}

\usepackage{amsmath,amsfonts,bm}

\def\eqref#1{equation~\ref{#1}}

\def\1{\bm{1}}

\DeclareMathAlphabet{\mathsfit}{\encodingdefault}{\sfdefault}{m}{sl}
\SetMathAlphabet{\mathsfit}{bold}{\encodingdefault}{\sfdefault}{bx}{n}

\usepackage{hyperref}
\usepackage{url}

\usepackage[utf8]{inputenc} 
\usepackage[T1]{fontenc}    
\usepackage{hyperref}       
\usepackage{url}            
\usepackage{booktabs}       
\usepackage{amsfonts}       
\usepackage{nicefrac}       
\usepackage{microtype}      
\usepackage{xcolor}         

\usepackage{amsmath}
\usepackage{pgfplots}
\usepackage{caption}
\usepackage{subcaption}
\usetikzlibrary{pgfplots.groupplots}
\usepackage{algorithm}
\usepackage{algorithmic}
\usepackage{booktabs}
\usepackage{multirow}
\usepackage{subcaption}  

\title{Context Attribution with Multi-Armed Bandit Optimization\thanks{Published as a Findings paper at ACL 2026.}}

\author{
Deng Pan$^1$\quad
Keerthiram Murugesan$^3$\quad
Ting Hua$^1$\quad
Nuno Moniz$^1$\quad
Nitesh V. Chawla$^{1,2}$ \\
$^1$Lucy Family Institute for Data \& Society, University of Notre Dame \\
$^2$Department of Computer Science \& Engineering, University of Notre Dame \\
$^3$IBM Research \\
\texttt{\{dpan, thua, nuno.moniz, nchawla\}@nd.edu} \\
\texttt{Keerthiram.Murugesan@ibm.com}
}

\pgfplotsset{compat=1.18}
\begin{document}

\maketitle

\begin{abstract}
Understanding which parts of the retrieved context contribute to a large language model's generated answer is essential for building interpretable and trustworthy retrieval-augmented generation. We propose a novel framework that formulates context attribution as a combinatorial multi-armed bandit problem. We utilize Linear Thompson Sampling to efficiently identify the most influential context segments while minimizing the number of model queries.  Our reward function leverages token log-probabilities to measure how well a subset of segments supports the original response, making it applicable to both open-source and black-box API-based models. Unlike SHAP and other perturbation-based methods that sample subsets uniformly, our approach adaptively prioritizes informative subsets based on posterior estimates of segment relevance, reducing computational costs. Experiments on multiple QA benchmarks demonstrate that our method achieves up to 30\% reduction in model queries while matching or exceeding the attribution quality of existing approaches. Our code is publicly available at \url{https://github.com/pd90506/camab}.
\end{abstract}
\section{Introduction}

Retrieval-Augmented Generation (RAG) has become a dominant approach for knowledge-intensive question answering tasks, augmenting Large Language Models (LLMs) with external context to improve factual accuracy and credibility \cite{gao2023retrieval}. Despite its effectiveness, ensuring that generated answers are genuinely grounded in the provided context remains challenging. LLMs frequently produce hallucinations or incorporate ungrounded information, making it essential to verify and attribute precisely which context segments are responsible for their responses \cite{gao2023enabling}.

Existing approaches to enhancing attribution primarily follow two paradigms. The first involves training models to explicitly cite context segments during generation \cite{nakano2021webgpt,menick2022teaching,zhang2024longcite,huang2024training}.  While such techniques improve self-attribution, their reliability remains \textbf{unverifiable}, as there is no guarantee that the generated citations actually reflect the context relied upon by the model during inference. The second paradigm focuses on post-hoc methods, such as ContextCite \cite{cohen2024contextcite} and MExGen \cite{paes2024multi}, which systematically perturb or mask context segments and evaluate their impact on the output. 
Traditional methods like LIME \cite{ribeiro2016should} and SHAP \cite{lundberg2017unified} also fall into this category with proper adaptation. Although these methods offer greater fidelity by probing actual input-output behavior, they often incur prohibitive computational costs due to extensive sampling and expensive LLM inference, rendering them impractical for long-context scenarios.

To address these limitations, we formulate context attribution as a Combinatorial Multi-Armed Bandit (CMAB) problem. Each context segment is treated as an arm, and selecting a subset of segments constitutes an action. The goal is to identify the most influential segments within a limited query budget.
To efficiently navigate this combinatorial space, we introduce Combinatorial \textbf{ Linear Thompson Sampling (LinTS)}, a Bayesian bandit method renowned for balancing exploration and exploitation \cite{agrawal2013thompson}.
Unlike exhaustive or uniformly random perturbation strategies, our approach significantly reduces the number of model evaluations, making it practical for long-context applications.

Our key contributions are summarized as follows: 1) we formulate segment-level context attribution as a combinatorial multi-armed bandit (CMAB) problem; 2) we utilize Linear Thompson Sampling (LinTS) to adaptively explore the space of context subsets, improving query efficiency under limited budgets; and 3) we conduct comprehensive experiments on three diverse datasets using two widely-used LLMs accessed via commercial APIs, demonstrating the effectiveness of our approach in realistic black-box settings.



\section{Proposed Method} \label{sec:proposed}

In this section, we present \textbf{CAMAB} (Context Attribution with Multi-Armed Bandit), a framework that treats segment-level attribution as an efficient Bayesian optimization process.
Rather than proposing a new bandit algorithm, our contribution is a \textit{problem reformulation} built on three non-trivial design decisions: (1) a \textbf{reward formulation} that measures joint subset contribution via the token-level log-probability of the original response (Eq.~\ref{eq:reward_linear}), providing a structured reward signal over combinatorial inputs; (2) a \textbf{combinatorial super-arm representation} using binary inclusion vectors (Eq.~\ref{eq:feature_vector}) that lets Thompson Sampling operate over the exponential action space of segment subsets; and (3) the use of the \textbf{precision matrix} $\boldsymbol{B}_t$ as an implicit interaction signal, whose off-diagonal entries accumulate co-selection statistics and capture dependencies between segments without requiring explicit non-linear modeling.
CAMAB leverages \textbf{Linear Thompson Sampling (LinTS)} to model the joint contribution of context subsets, achieving querying efficiency by exploring the context subset space adaptively.

\subsection{Problem Formulation}

\paragraph{Context-Supported Generative QA}
We consider a scenario where an LLM is tasked with answering a question $Q$ using a provided context $C$. The context $C$ consists of $N$ discrete segments, $C = \{s_1, s_2, \dots, s_N\}$, such as sentences or paragraphs. The LLM (denoted by $M$) produces a response $R$ consisting of a sequence of tokens $R = (r_1, r_2, \dots, r_T)$, generated autoregressively based on the input $(Q, C)$. While our formulation operates at the segment (sentence) level, the framework is granularity-agnostic: segments can be defined at any level of abstraction, from individual tokens to entire paragraphs. We adopt sentence-level segmentation as a practical compromise, as finer granularities (e.g., tokens) would exponentially expand the combinatorial action space, while coarser granularities (e.g., paragraphs) may obscure fine-grained attribution signals.

\paragraph{Reward Modeling via Context Subsets}
The central insight is to gauge a segment’s importance by observing the model’s output probability when that segment is included in or excluded from various subsets. Let $S \subseteq C$ denote a subset of the context segments. We denote by $P_M(\cdot)$ the conditional next-token probability induced by the language model $M$. For any subset $S$, we define the reward $V(S)$ as the average log-probability of the original response tokens $R$ given $S$:
\begin{equation}
\label{eq:reward_linear}
    V(S) = \frac{1}{T}\sum_{t=1}^T \log P_M(r_t \mid Q, S, r_1, \ldots, r_{t-1})
\end{equation}
This reward formulation serves as our metric for segment importance. While alternative definitions exist as noted by \cite{paes2024multi}, for simplicity and consistency, we adopt Eq~\ref{eq:reward_linear} as the standard reward throughout this work unless explicitly stated otherwise.

\paragraph{Context Attribution}
Given the reward modeling, our goal is to attribute the content of model response $R$ back to specific segments in context $C$ by assigning an attribution vector $\boldsymbol{a}$, where each score $a_j$ reflects the marginal contribution of segment $s_j$ to the generation of $R$.

\begin{table*}[htbp]
\centering
\caption{
BERTScore performance (lower is better) of \textbf{CAMAB}, \textbf{ContextCite}, and \textbf{SHAP} under varying query budgets ($s=20,40,60$) across datasets using the LLaMA3.1-8B model. Each $s$ denotes the total number of LLM calls used. Bold indicates the best performance.
}
\resizebox{0.85\textwidth}{!}{%
\begin{tabular}{llccccccccc}
\toprule
\textbf{Data} & \textbf{$k$} 
& \multicolumn{3}{c}{\textbf{CAMAB}} 
& \multicolumn{3}{c}{\textbf{ContextCite}} 
& \multicolumn{3}{c}{\textbf{SHAP}} \\
\cmidrule(lr){3-5} \cmidrule(lr){6-8} \cmidrule(lr){9-11}
& & $s=20$ & $s=40$ & $s=60$ & $s=20$ & $s=40$ & $s=60$ & $s=20$ & $s=40$ & $s=60$ \\
\midrule

\multirow{3}{*}{\shortstack{Hotpot\\QA}} 
    & 1 & 0.525 & \textbf{0.509} & 0.511 & 0.605& 0.601& 0.598& 0.668 & 0.562 & 0.527\\
    & 3 & 0.464 & 0.421 & \textbf{0.418} & 0.549& 0.537& 0.529& 0.598 & 0.471 & 0.444\\
    & 5 & 0.445 & 0.407 & \textbf{0.402} & 0.527& 0.496& 0.499& 0.562 & 0.453 & 0.423\\

\midrule

\multirow{3}{*}{\shortstack{CNN/\\DM}}
    & 1 & 0.642& \textbf{0.613} & \underline{0.614}& 0.671& 0.668& 0.670& 0.648& 0.617 & \textbf{0.613}\\
    & 3 & 0.526& 0.485& \underline{0.480}& 0.544& 0.535& 0.532& 0.532& 0.483 & \textbf{0.478}\\
    & 5 & 0.452& \underline{0.405}& 0.406& 0.479& 0.459& 0.454& 0.455& 0.406 & \textbf{0.399}\\

\midrule

\multirow{3}{*}{\shortstack{TyDi\\QA}} 
    & 1 & 0.479& \underline{0.473}& \textbf{0.470}& 0.542& 0.539& 0.536& 0.542& 0.488 & 0.476\\
    & 3 & 0.392& \textbf{0.378} & \underline{0.381}& 0.474& 0.468& 0.458& 0.455& 0.400 & 0.379\\
    & 5 & 0.371& \underline{0.345}& \textbf{0.343}& 0.452& 0.442& 0.440& 0.420& 0.368 & 0.353\\
\bottomrule
\end{tabular}
}
\label{tab:camab_contextcite_shap_results}
\end{table*}

\subsection{Bandit Formulation with Linear Thompson Sampling}

\paragraph{Combinatorial Multi-Armed Bandit}
We formulate context attribution as a Combinatorial Multi-Armed Bandit (CMAB) problem. In this setting, the $N$ context segments $\{s_1, \dots, s_N\}$ serve as the base arms of the bandit. An action corresponds to selecting a subset of segments $S \subseteq C$ (referred to as a \textit{super-arm}), and the observed feedback is the scalar reward $V(S)$, which reflects the model's response quality given the subset $S$.

To efficiently estimate segment importance, we adopt a linear structural assumption. We posit that the reward $V(S)$ can be approximated as the sum of the marginal contributions of the individual segments included in $S$, plus a bias term representing the model's baseline performance given an empty context. 
Specifically, for a selected subset $S$, we define a feature vector $\boldsymbol{x} \in \{0, 1\}^{N+1}$ to represent the combinatorial action:
\begin{equation}
\label{eq:feature_vector}
    \boldsymbol{x} = [1, \mathbb{I}(s_1 \in S), \dots, \mathbb{I}(s_N \in S)]^\top
\end{equation}
where the first element is a bias term (intercept) representing the model's performance with an empty context, and $\mathbb{I}(\cdot)$ is the indicator function for segment inclusion. We model the reward as a linear function of the features with additive Gaussian noise:
\begin{align}
    V(S) = \boldsymbol{w}^\top \boldsymbol{x} + \epsilon, 
\end{align}
where $\epsilon \sim \mathcal{N}(0, \sigma^2)$ is the observation noise and $\boldsymbol{w} = [w_0, w_1, \dots, w_N]^\top$ represents the latent weights. In this formulation, $w_0$ captures the baseline log-likelihood, and $w_j$ for $j > 0$ denotes the specific contribution of segment $s_j$. The final learned weights $\{w_1, \dots, w_N\}$ serve as our attribution score vector $\boldsymbol{a}$.

\paragraph{Linear Thompson Sampling}
To solve this problem under a strictly limited query budget, we employ Linear Thompson Sampling (LinTS)~\cite{agrawal2013thompson}. LinTS is a sample-efficient algorithm that balances exploration and exploitation by maintaining a probabilistic belief over the unknown weight vector $\boldsymbol{w}$. Assuming a Gaussian prior $\boldsymbol{w} \sim \mathcal{N}(\hat{\boldsymbol{\mu}}_0, \boldsymbol{B}_0^{-1})$ and Gaussian observation noise $\epsilon \sim \mathcal{N}(0, \sigma^2)$, the posterior distribution at any round $t$ remains a multivariate Gaussian due to conjugacy:
\begin{equation}
    P(\boldsymbol{w} \mid \mathcal{H}_t) = \mathcal{N}(\hat{\boldsymbol{\mu}}_t, \boldsymbol{B}_t^{-1})
\end{equation}
where $\mathcal{H}_t$ denotes the history of observed actions and rewards up to round $t$. The algorithm proceeds by iteratively sampling a plausible weight vector $\tilde{\boldsymbol{w}}^{(t)}$ from this posterior to guide the selection of the next informative context subset.

Furthermore, although we rely on a linear assumption, \textit{the precision matrix $\boldsymbol{B}_t$ allows the model to implicitly capture interactions}. By tracking correlations between segments, $\boldsymbol{B}_t$ ensures that if a subset of segments consistently performs well together, this interaction effect is absorbed and reflected in their respective individual weights.

\begin{algorithm}[t]
\caption{CAMAB: Context Attribution with Linear Thompson Sampling}
\label{alg:camab}
\begin{algorithmic}[1]
\REQUIRE Context segments $C = \{s_1, \dots, s_N\}$, query $Q$, response $R$, budget $T_{\max}$, prior mean $\hat{\boldsymbol{\mu}}_0$, prior variance $\sigma_p^2$, noise variance $\sigma^2$
\ENSURE Attribution scores $\{w_1, \dots, w_N\}$
\STATE \textbf{Initialize:} $\boldsymbol{B}_0 \leftarrow \frac{1}{\sigma_{p}^2} \boldsymbol{I}$, $\boldsymbol{f}_0 \leftarrow \boldsymbol{B}_0 \hat{\boldsymbol{\mu}}_0$
\FOR{$t = 1$ to $T_{\max}$}
    \STATE Sample $\tilde{\boldsymbol{w}}^{(t)} \sim \mathcal{N}\left(\hat{\boldsymbol{\mu}}_{t-1}, \boldsymbol{B}_{t-1}^{-1}\right)$
    \STATE Construct $\boldsymbol{x}_t$ and form subset $S_t$
    \STATE Query LLM and observe reward $v_t = V(S_t)$ via Eq.~\ref{eq:reward_linear}
    \STATE Update $\boldsymbol{B}_t \leftarrow \boldsymbol{B}_{t-1} + \frac{1}{\sigma^2} \boldsymbol{x}_t \boldsymbol{x}_t^\top$
    \STATE Update $\boldsymbol{f}_t \leftarrow \boldsymbol{f}_{t-1} + \frac{v_t}{\sigma^2} \boldsymbol{x}_t$
    \STATE Compute $\hat{\boldsymbol{\mu}}_t \leftarrow \boldsymbol{B}_t^{-1} \boldsymbol{f}_t$
\ENDFOR
\RETURN $\{\hat{\mu}_1, \dots, \hat{\mu}_N\}$ from $\hat{\boldsymbol{\mu}}_{T_{\max}}$
\end{algorithmic}
\end{algorithm}

Algorithm~\ref{alg:camab} summarizes our approach. At each round $t$, we first sample a weight vector $\tilde{\boldsymbol{w}}^{(t)}$ from the current posterior distribution (Line 3). This sample reflects our current belief about segment contributions while incorporating uncertainty. We then construct the context subset $S_t$ by selecting all segments whose sampled weights are positive (Line 4), effectively choosing segments that are likely to contribute positively to the response:
\begin{equation}
    S_t = \{s_j : \tilde{w}_j^{(t)} > 0\}
\end{equation}
The corresponding feature vector is constructed as:
\begin{equation}
    \boldsymbol{x}_t = [1, \mathbb{I}(\tilde{w}_1^{(t)} > 0), \dots, \mathbb{I}(\tilde{w}_N^{(t)} > 0)]^\top
\end{equation}
After querying the LLM with subset $S_t$ to observe the reward $v_t$ (Line 5), we perform a Bayesian update: the precision matrix $\boldsymbol{B}_t$ accumulates information about which segment combinations have been tested, while the information vector $\boldsymbol{f}_t$ aggregates the observed rewards weighted by the corresponding feature vectors (Lines 6-8). This posterior update naturally balances exploration of uncertain segments with exploitation of segments already identified as important.

After $T_{\max}$ rounds, the posterior mean weights $\hat{\mu}_1, \dots, \hat{\mu}_N$ (excluding the intercept $\hat{\mu}_0$) are used as the final attribution scores. Segments are ranked by these scores to identify the most influential context components.



\section{Experiments} \label{sec:experiments}

We evaluate CAMAB against \textbf{SHAP}~\cite{lundberg2017unified}, \textbf{ContextCite}~\cite{cohen2024contextcite}, and a random baseline on three diverse benchmarks---HotpotQA~\cite{yang2018hotpotqa}, CNN/DailyMail~\cite{see2017get}, and TyDi QA~\cite{clark2020tydi}---using a random subset of 10,000 samples from each. 

Notably, ContextCite's official implementation relies on full logit-based reward, while we set CAMAB and SHAP operate solely on token-level log-probabilities (Eq~\ref{eq:reward_linear}).
We employ two state-of-the-art models: LLaMA-3.1-8B~\cite{grattafiori2024llama} and Qwen2.5-7B~\cite{qwen2025qwen25technicalreport}. Performance is assessed using the mean value of \textbf{Log-Probability Drop} (higher is better $\uparrow$) and \textbf{BERTScore} (lower is better $\downarrow$) over the 10,000 samples. Detailed experimental settings are provided in Appendix~\ref{sec:additional}.

\subsection{Attribution Quality on Different Tasks}

Table~\ref{tab:eval_llama} presents attribution performance under a tight budget of 40 queries. We observe that CAMAB consistently outperforms or matches the baselines across all datasets. Specifically, on information-seeking tasks such as HotpotQA and TyDi QA, CAMAB achieves superior performance in both Top-$k$ Log-Probability Drop and BERTScore metrics for nearly all $k$ values. This demonstrates that our bandit-based approach is highly effective at pinpointing the critical evidence sentences required for factual reasoning.

On the abstractive summarization task (CNN/DailyMail), CAMAB remains highly competitive, achieving performance almost identical to SHAP with a difference of less than 1\% across all metrics. We attribute this to the strong \textit{lead bias} in news summarization, where key information is concentrated at the beginning and end of articles. This structure reduces ambiguity, allowing different attribution methods to converge rapidly. This hypothesis is corroborated by Table~\ref{tab:camab_contextcite_shap_results}, where all methods stabilize within 40 queries.

Consistent trends are observed for the Qwen-2.5-8B model; detailed results are provided in Appendix~\ref{sec:additional}.

\begin{table}[htbp]
\centering
\caption{Evaluation results on LLaMA-3.1-8B with querying budget of 40. Abbreviations are used for conciseness (C-Cite: ContextCite, Rand: Random, Log-P: Log-Probability Drop, BERT: BERTScore).}
\label{tab:eval_llama}
\setlength{\tabcolsep}{3pt}
\resizebox{0.9\linewidth}{!}{%
\begin{tabular}{llccccc}
\toprule
\textbf{Data} & \textbf{Metric} & \textbf{$k$} & \textbf{CAMAB} & \textbf{SHAP} & \textbf{C-Cite} & \textbf{Rand} \\
\midrule

\multirow{6}{*}{\shortstack{Hotpot\\QA}} & \multirow{3}{*}{{Log-P $\uparrow$}} & 1 & \textbf{0.521} & \underline{0.475} & 0.429 & 0.024\\
 & & 3 & \textbf{0.676} & \underline{0.614} & 0.591 & 0.062\\
 & & 5 & \textbf{0.717} & \underline{0.648} & 0.632 & 0.103\\
\cmidrule(lr){2-7}
 & \multirow{3}{*}{{BERT $\downarrow$}} & 1 & \textbf{0.509} & \underline{0.562} & 0.601 & 0.803\\
 & & 3 & \textbf{0.421} & \underline{0.471} & 0.537 & 0.741\\
 & & 5 & \textbf{0.407} & \underline{0.453} & 0.496 & 0.703\\
\midrule

\multirow{6}{*}{\shortstack{CNN/\\DM}} & \multirow{3}{*}{{Log-P $\uparrow$}} & 1 & \textbf{0.400} & \underline{0.398} & 0.358 & 0.073\\
 & & 3 & \underline{0.840} & \textbf{0.843} & 0.801 & 0.224\\
 & & 5 & \textbf{1.129} & \underline{1.041} & 1.025 & 0.389\\
\cmidrule(lr){2-7}
 & \multirow{3}{*}{{BERT $\downarrow$}} & 1 & \textbf{0.613} & \underline{0.617} & 0.668 & 0.734\\
 & & 3 & \underline{0.485} & \textbf{0.483} & 0.535 & 0.656\\
 & & 5 & \textbf{0.405} & \underline{0.406} & 0.459 & 0.601\\
\midrule

\multirow{6}{*}{\shortstack{TyDi\\QA}} & \multirow{3}{*}{{Log-P $\uparrow$}} & 1 & \textbf{0.596} & \textbf{0.596} & \underline{0.429} & 0.069\\
 & & 3 & \textbf{0.813} & \underline{0.803} & 0.579 & 0.240\\
 & & 5 & \textbf{0.893} & \underline{0.872} & 0.631 & 0.373\\
\cmidrule(lr){2-7}
 & \multirow{3}{*}{{BERT $\downarrow$}} & 1 & \textbf{0.473} & \underline{0.488} & 0.539& 0.719\\
 & & 3 & \textbf{0.378} & \underline{0.400} & 0.468& 0.624\\
 & & 5 & \textbf{0.345} & \underline{0.368} & 0.442& 0.566\\
\bottomrule
\end{tabular}
}
\end{table}

\subsection{Attribution with Limited Query Budgets}

In realistic scenarios, query budgets are strictly constrained due to the high latency and monetary cost of LLM inference. To evaluate the sample efficiency of our framework, we compare CAMAB against ContextCite and SHAP across three query budgets: $s \in \{20, 40, 60\}$. Table~\ref{tab:camab_contextcite_shap_results} reports the BERTScore (lower is better) on LLaMA-3.1-8B.

We observe three key trends. \textbf{First, CAMAB demonstrates superior sample efficiency.} In information-seeking tasks (HotpotQA and TyDi QA), CAMAB at $s=40$ often outperforms SHAP at $s=60$.  This indicates that the active exploration of LinTS identifies critical segments significantly faster than random perturbations.
\textbf{Second, the performance gap narrows as the query budget increases.} While SHAP eventually converges to competitive results when sufficient queries are available (e.g., $s=60$), CAMAB achieves near-optimal performance significantly earlier. At $s=20$, SHAP's performance degrades sharply, whereas CAMAB maintains high fidelity even in these extremely resource-constrained settings.
\textbf{Finally, CAMAB consistently outperforms ContextCite.} ContextCite struggles to identify influential segments under tight budgets, likely because Lasso regression requires a larger sample size to effectively select features in high-dimensional spaces.

These results validate that CAMAB is the most robust choice for attribution when computational resources are limited. We further verify that these trends hold at $s{=}100$; see Appendix~\ref{sec:larger_budget} for details. We also verify that CAMAB's $O(N^3)$ posterior update adds negligible wall-clock overhead compared to LLM inference; see Appendix~\ref{sec:wallclock} for details.

\subsection{Alignment with Gold Supporting Facts}

Beyond perturbation-based metrics, we evaluate whether CAMAB's attributed segments agree with human-annotated supporting facts. We construct a 200-sample set from HotpotQA \texttt{distractor/validation} (each with $\leq$40 segments and $\geq$1 gold fact) and rank segments by attribution scores at $s{=}40$. We report Precision at 1 (P@1), F1 at $k{=}2$ and at oracle $k^*{=}|\text{gold}|$, Area Under the ROC Curve (AUROC), and Average Precision (AP), where AP summarizes the area under the Precision--Recall curve.

\begin{table}[t]
\centering
\caption{Alignment with HotpotQA gold supporting facts (200 samples,
LLaMA-3.1-8B, $s{=}40$). $k^*$: oracle $k$ = $|\text{gold}|$.}
\label{tab:supporting_facts}
\small
\begin{tabular}{lccccc}
\toprule
\textbf{Method} & \textbf{P@1} & \textbf{F1@2} & \textbf{F1@k*} & \textbf{AUROC} & \textbf{AP} \\
\midrule
CAMAB  & \textbf{0.780} & \textbf{0.607} & \textbf{0.616} & \textbf{0.855} & \textbf{0.688} \\
SHAP   & 0.680 & 0.511 & 0.513 & 0.806 & 0.598 \\
Random & 0.055 & 0.058 & 0.071 & 0.516 & 0.162 \\
\bottomrule
\end{tabular}
\end{table}

As shown in Table~\ref{tab:supporting_facts}, CAMAB consistently outperforms SHAP across all metrics, confirming that our method effectively identifies the most informative context segments and produces attributions that agree with human judgment.

\subsection{Interaction Effects in the Precision Matrix}

A potential concern with the linear reward assumption (Eq.~\ref{eq:reward_linear}) is that it may miss synergies between segments. We show empirically that the precision matrix $\boldsymbol{B}_t$ captures such interactions implicitly: posterior correlations cluster within sub-topics, revealing substitutability that marginal scores alone cannot express. A detailed case study with visualization is provided in Appendix~\ref{sec:interaction}.

\section{Conclusion}
We introduced CAMAB, a novel context attribution framework that formulates attribution as a combinatorial multi-armed bandit problem and applies Linear Thompson Sampling to efficiently attribute the contexts. 
Empirical results demonstrate that CAMAB delivers superior attribution fidelity with significantly fewer queries, establishing it as a scalable solution for low-resource generative QA systems.

\newpage

\section*{Limitations}
While CAMAB demonstrates strong performance under constrained query budgets, it is primarily designed for scenarios where query efficiency is critical, such as black-box attribution for large generative language models. However, this approach may converge to suboptimal local solutions if the exploration-exploitation balance is not well maintained, especially in highly noisy or ambiguous settings. In situations where budget is not a constraint, traditional perturbation-based methods such as SHAP and ContextCite can benefit from broader context exploration and may ultimately yield more plausible attributions. Additionally, CAMAB requires access to token-level log-probabilities, which are available through many commercial APIs (e.g., OpenAI, Together) but not universally exposed by all providers; extending the framework to settings where only generated text is observable remains an open direction.

\paragraph{Use of AI Assistants} We utilized AI assistants for grammatical error correction, text polishing, and code assistance. The authors verified all generated content for accuracy.

\bibliography{ref}

\appendix
\section*{Appendix}

\section{Related Work}

\subsection{Perturbation-Based Attribution Methods}
A large number of works on model interpretability uses input perturbations to infer feature importance. Techniques like \textbf{LIME} \cite{ribeiro2016should} and \textbf{SHAP} \cite{lundberg2017unified} interpret model predictions by evaluating the model on perturbations of an input and observing how the output changes. LIME fits a local surrogate model (e.g. a linear model) around the neighborhood of the input to estimate each feature’s influence. SHAP uses a game-theoretic approach to approximate Shapley values, which quantify each feature’s contribution to the prediction in a way that satisfies fairness axioms. These methods are \emph{model-agnostic} and fairly \textit{faithful} in a local sense, but they are notoriously expensive: they require sampling a large number of perturbations for each instance to obtain stable estimates. This cost grows with input dimensionality, making them difficult to apply to settings like long text sequences without sparing accuracy. 

\subsection{Attribution Methods in LLM Settings}
In the context of large language models (LLMs), token-level attribution faces significant challenges due to (1) the combinatorially large perturbation space induced by extensive input contexts, and (2) substantial computational costs associated with individual model queries. Therefore, effective attribution methods must carefully balance explanation fidelity against computational efficiency. Broadly, three strategies have emerged to tackle these challenges:

(i) \textbf{Reducing Perturbation Space}: Methods in this category aggregate tokens into higher-level semantic units, such as phrases, sentences, or paragraphs, effectively decreasing the perturbation space. Different granularity levels inherently capture varying degrees of semantic meaning, naturally supporting hierarchical attribution structures. For example, \cite{chen2020generating} propose a divide-and-conquer strategy that progressively attributes importance from sentence-level down to token-level in text classification. Similarly, MExGen~\cite{paes2024multi} systematically extends perturbation-based methods like LIME and SHAP to generative LLMs, efficiently identifying influential text spans in a hierarchical manner. ContextCite~\cite{cohen2024contextcite} specifically targets segment-level attribution in generative QA scenarios by using SHAP-based perturbation techniques.

(ii) \textbf{Pretrained Global Explainers}: Another strategy involves training a global surrogate model as a pretrained explainer, trading upfront training costs for reduced inference latency during explanations. Examples include FEX~\cite{pan2025fast}, which employs policy gradient methods to optimize a Bernoulli surrogate explainer, and FastSHAP~\cite{jethani2021fastshap}, which fits a neural network using pseudo-labels derived from SHAP values. Despite their inference efficiency, these methods demand substantial pretraining resources and extensive datasets, and like typical machine learning models, they often encounter generalization challenges when confronted with out-of-distribution samples.

(iii) \textbf{Optimizing Perturbation Strategies}: A relatively nascent direction focuses on explicitly optimizing perturbation strategies to reduce the number of required model queries. \cite{sudhakar2021ada} leverage heuristics derived from input-to-output gradients to selectively perturb features, whereas \cite{pan2021explaining} propose subsequent perturbations aligned with adversarial attack directions. However, such methods typically require internal model knowledge, including gradients or manifold structures, limiting their applicability to models treated as black boxes. Motivated by this gap, our work introduces a novel perturbation sampling method inspired by multi-armed bandit algorithms, enabling dynamic adjustment of perturbations based solely on observed responses, without necessitating internal model information.


\subsection{Bandit and Reinforcement Learning Approaches} 
Feature attribution can be viewed as a task of selecting the most informative subsets of features. The multi-armed bandit and reinforcement learning can be utilized to progressively optimize and sequentially search the subsets with a limited budget of actions. Feature selection via multi-armed bandits has been explored in prior research as a way to dynamically identify important features without evaluating all subsets. For example, \cite{nagaraju2025automation} propose a feature selection method that uses an Upper Confidence Bound (UCB) bandit algorithm. This allows the algorithm to rapidly converge to a near-optimal feature set, yielding good predictors with fewer feature evaluations. In NLP tasks, BanditSum \cite{dong2018banditsum} treated extractive summarization as a contextual bandit problem: given a document (context), their model learned via policy gradient to pick a sequence of sentences (the “action”) that maximizes the summary quality reward (ROUGE score) This is an example of using reinforcement learning to select informative subsets of text. Although BanditSum was focused on training a summarization model, the idea of using reward feedback to guide text segment selection is closely related to our approach for attribution.

\section{Additional Experiment Details and Results} \label{sec:additional}

\subsection{Datasets}

We evaluate our framework on three representative language generation benchmarks that cover distinct task types and context structures. HotpotQA targets sentence-level attribution in multi-hop question answering; CNN/DailyMail emphasizes sentence-level attribution for long-document summarization; and TyDi QA provides a diverse, multilingual context for evaluating information-seeking question answering. For computational feasibility, we randomly sample 1,000 validation instances from each dataset.

\paragraph{HotpotQA}~\cite{yang2018hotpotqa}
HotpotQA is a multi-hop question answering benchmark requiring reasoning over multiple supporting documents to answer factoid questions. Each instance includes long passages, and the responses are more elaborate than standard QA tasks. We use sentences as the segments of interest for attribution.

\paragraph{CNN/DailyMail}~\cite{see2017get}
CNN/DailyMail is a large-scale abstractive summarization dataset where the task is to generate concise summaries of news articles. Contexts consist of long documents containing narrative and factual information, and the outputs are multi-sentence summaries. We select sentences as the segments of interest.

\paragraph{TyDi QA}~\cite{clark2020tydi}
TyDi QA is a benchmark for information-seeking question answering, grounded in Wikipedia contexts across multiple typologically diverse languages. Unlike the multi-hop reasoning in HotpotQA, TyDi QA focuses on identifying specific evidence from a provided context to answer user questions. We treat individual sentences as the segments of interest to evaluate the precision of our attribution framework in factual retrieval scenarios.

\subsection{Models}

To evaluate the generality and robustness of our context attribution framework, we conduct experiments with two recent large language models that differ in their training corpora and design philosophies. To reflect a realistic black-box attribution scenario, these models are accessed via commercial APIs, where internal model states and full logit distributions are unavailable, and only token-level log-probabilities are provided.

\paragraph{LLaMA-3.1-8B.}
We use the 8B version of LLaMA 3.1~\cite{grattafiori2024llama}, a decoder-only transformer released by Meta AI. Pretrained on a vast and diverse corpus with a next-token prediction objective, LLaMA-3.1 excels at long-context reasoning and producing fluent, high-quality responses. In our experiments, it serves as a high-capacity baseline to test attribution fidelity in complex reasoning tasks.

\paragraph{Qwen2.5-7B.}
We include Qwen2.5-7B~\cite{qwen2025qwen25technicalreport}, a decoder-only transformer developed by Alibaba. Trained on large-scale multilingual and multimodal data, Qwen2.5 demonstrates strong cross-lingual generalization and represents a distinct pretraining paradigm from LLaMA-3.1. Including this model allows us to evaluate the consistency of our attribution framework across different architectural families and training objectives.

Overall, these two models provide a complementary testbed for evaluating CAMAB in a black-box setting, representing high-capacity systems commonly deployed in real-world knowledge-intensive applications.

\begin{table*}[htbp]
\centering
\caption{Evaluation results on Qwen2.5-7B. Top-$k$ means the specific evaluation is done with top $k$ attributed segments removed.}
\label{tab:eval_qwen}
\resizebox{0.7\linewidth}{!}{%
\begin{tabular}{lllcccc}
\toprule
\textbf{Dataset} & \textbf{Metrics} & \textbf{Top-$k$} & \textbf{CAMAB} & \textbf{SHAP} & \textbf{ContextCite} & \textbf{Random} \\
\midrule

\multirow{6}{*}{HotpotQA} & \multirow{3}{*}{Log-Prob Drop \textuparrow} & $k=1$ & \textbf{0.650} & \underline{0.628} & 0.541 & 0.041\\
 & & $k=3$ & \textbf{0.917} & \underline{0.879} & 0.802 & 0.107\\
 & & $k=5$ & \textbf{0.972} & \underline{0.948} & 0.867 & 0.184\\
\cmidrule(lr){2-7}
 & \multirow{3}{*}{BERTScore \textdownarrow} & $k=1$ & \textbf{0.486} & 0.504 & \underline{0.490} & 0.782\\
 & & $k=3$ & \textbf{0.372} & 0.394 & \underline{0.393} & 0.704\\
 & & $k=5$ & \textbf{0.355} & 0.375 & \underline{0.370} & 0.649\\
\midrule
\multirow{6}{*}{CNN/DailyMail} & \multirow{3}{*}{Log-Prob Drop \textuparrow} & $k=1$ & \textbf{0.442} & \underline{0.393} & 0.365 & 0.112\\
 & & $k=3$ & \textbf{0.998} & \underline{0.910} & 0.837 & 0.330\\
 & & $k=5$ & \textbf{1.371} & \underline{1.285} & 1.192 & 0.570\\
\cmidrule(lr){2-7}
 & \multirow{3}{*}{BERTScore \textdownarrow} & $k=1$ & \textbf{0.613} & \underline{0.624} & 0.632 & 0.712\\
 & & $k=3$ & \textbf{0.487} & \underline{0.504} & 0.542 & 0.631\\
 & & $k=5$ & \textbf{0.415} & \underline{0.430} & 0.478 & 0.570\\
\midrule
\multirow{6}{*}{TyDi QA} & \multirow{3}{*}{Log-Prob Drop \textuparrow} & $k=1$ & \underline{0.732} & \textbf{0.738} & 0.496 & 0.107\\
 & & $k=3$ & \textbf{0.994} & \underline{0.992} & 0.719 & 0.269\\
 & & $k=5$ & \textbf{1.081} & \underline{1.072} & 0.792 & 0.436\\
\cmidrule(lr){2-7}
 & \multirow{3}{*}{BERTScore \textdownarrow} & $k=1$ & \textbf{0.430} & \underline{0.448} & 0.509 & 0.694\\
 & & $k=3$ & \textbf{0.348} & \underline{0.362} & 0.443 & 0.605\\
 & & $k=5$ & \textbf{0.325} & \underline{0.347} & 0.423 & 0.541\\
\bottomrule
\end{tabular}
}
\end{table*}

\subsection{Settings and Baselines}

We compare CAMAB against four representative post-hoc attribution baselines, each reflecting a different strategy for identifying influential context segments. To ensure a fair comparison, all methods are evaluated under a strictly constrained query budget of $T=40$ LLM calls per instance.

\paragraph{CAMAB (Ours)}
We implement CAMAB using Linear Thompson Sampling (LinTS) as described in Section~\ref{sec:proposed}. The algorithm is initialized with a prior mean of $0.0$ and a prior variance of $1.0$. The observation noise variance is set to $\sigma^2 = 0.01$. Our implementation employs an \textbf{adaptive selection} strategy: in each round, it selects all segments whose sampled weights $\tilde{\theta}_j$ are positive. This allows the model to dynamically adjust the size of the context subset based on its current posterior beliefs.

\paragraph{SHAP}
SHAP~\cite{lundberg2017unified} is a model-agnostic explainer grounded in Shapley values. We use the KernelSHAP variant at the segment level. To comply with the query budget, we use a single fully masked context as the reference baseline and limit the number of perturbed samples to $40$. The reward signal is the average log-likelihood of the response tokens, consistent with our method.

\paragraph{ContextCite}
ContextCite~\cite{cohen2024contextcite} attributes generation by measuring the average log-odds change when subsets of segments are ablated. It fits a sparse LASSO regression model to these observations to identify relevant segments. To ensure parity in computational cost, we limit its sampling to $40$ ablated subsets.


\paragraph{Random}
As a baseline for attribution effectiveness, the Random explainer assigns importance scores by sampling from a uniform distribution. This provides a lower-bound reference to demonstrate that the performance gains of other methods are due to principled exploration rather than random perturbations.

\subsection{Experiment Settings}
To simulate a realistic black-box scenario, our experiments for CAMAB and the baselines SHAP and Random are conducted using commercial APIs, where only token-level log-probabilities are accessible. Contextcite, on the otherhand, requires full access to the model's logit distributions to perform its regression-based attribution, hence we conduct its experiments on a local machine. 

This setup reflects the typical constraints of state-of-the-art models where internal states are unavailable. 

In contrast, ContextCite requires full access to the model's logit distributions to perform its regression-based attribution. Consequently, the experiments for ContextCite were conducted on a local computing server equipped with an NVIDIA A100 GPU (80GB). To ensure a fair comparison, all methods are evaluated under strictly constrained query budgets, controlling for the total number of LLM calls across all attribution processes. 

\subsection{Evaluation Metrics}

To assess the effectiveness of our context attribution method, we adopt two evaluation metrics: \emph{Top-$k$ Log-Probability Drop} and \emph{BERTScore Consistency}.

\paragraph{Top-$k$ Log-Probability Drop.} \cite{cohen2024contextcite}
This metric evaluates the degradation of the model response's likelihood when the most influential segments are removed. 
Let $L(S)$ denote the average log-likelihood of the response $R$ given context subset $S$:
\begin{equation}
    L(S) = \frac{1}{T} \sum_{t=1}^{T} \log P_{M}(r_t \mid Q, S, r_{<t})
\end{equation}
The Top-$k$ log-probability drop is then defined as the difference in likelihood between the full context $C$ and the perturbed subset $S_{\text{top-}k}(\tau)$:
\begin{equation}
    \text{Top-}k\text{-drop} = L(C) - L(S_{\text{top-}k}(\tau))
\end{equation}
A larger drop implies that the removed segments were more supportive of the generation, indicating higher attribution accuracy.

\paragraph{BERTScore Consistency.}
This metric evaluates attribution fidelity by measuring the semantic difference between the original response $R = (r_1, \dots, r_T)$, generated using the full context $C$, and the response $R'$, generated using the perturbed context $S_{\text{top-}k}(\tau)$. We compute the BERTScore~\citep{zhang2019bertscore} between the two responses as:
\begin{equation}
\text{BERTScore} = \text{BERTScore}\left(R', R\right)
\end{equation}
A lower BERTScore indicates a greater semantic shift caused by the ablation, suggesting that the removed segments were more influential. Thus, lower values reflect more accurate attribution.

\subsection{Wall-Clock Latency}
\label{sec:wallclock}

A potential concern is whether CAMAB's $O(N^3)$ posterior update (matrix inversion at each round) adds meaningful overhead compared to the LLM inference cost that dominates each query. Table~\ref{tab:wallclock} reports end-to-end wall-clock latency on 100 HotpotQA samples using the LLaMA-3.1-8B API with a budget of $s{=}100$. CAMAB is approximately 8\% faster than SHAP in total time per sample, as LinTS's adaptive subset selection tends to produce shorter context subsets (fewer tokens per query), which more than offsets the matrix algebra overhead.

\begin{table}[t]
\centering
\caption{Wall-clock latency comparison on HotpotQA (100 samples, LLaMA-3.1-8B, $s{=}100$). The $O(N^3)$ matrix update overhead in CAMAB is negligible relative to LLM inference cost; CAMAB is slightly faster overall due to fewer average tokens per query.}
\label{tab:wallclock}
\small
\begin{tabular}{lccc}
\toprule
\textbf{Method} & \textbf{Time (s/sample)} & \textbf{Avg.\ Segments} & \textbf{Avg.\ Tokens} \\
\midrule
CAMAB & 33.86 & 17.2 & 2{,}085 \\
SHAP  & 36.88 & 16.8 & 2{,}258 \\
\bottomrule
\end{tabular}
\end{table}

\subsection{Attribution with Larger Query Budget}
\label{sec:larger_budget}

To further examine CAMAB's behavior as the query budget increases, we evaluate on HotpotQA with $s{=}100$ using LLaMA-3.1-8B (1,000 samples). Table~\ref{tab:budget_100} reports both Log-Probability Drop and BERTScore alongside the $s \in \{20,40,60\}$ results for comparison.

CAMAB's performance saturates around $s{=}60$: increasing the budget to $s{=}100$ yields negligible improvement across all metrics. In contrast, SHAP requires $s{=}100$ to reach the level CAMAB achieves at $s{=}60$ (e.g., BERTScore@3: SHAP 0.440 at $s{=}100$ vs.\ CAMAB 0.418 at $s{=}60$). This confirms that CAMAB converges significantly faster than SHAP, validating its sample efficiency advantage even when computational constraints are relaxed.

\begin{table}[t]
\centering
\caption{HotpotQA attribution performance under varying query budgets ($s \in \{20, 40, 60, 100\}$) using LLaMA-3.1-8B (1,000 samples). Log-Probability Drop (higher $\uparrow$) and BERTScore (lower $\downarrow$). Bold indicates the best per $k$.}
\label{tab:budget_100}
\small
\begin{tabular}{llcccc}
\toprule
\textbf{Method} & \textbf{$k$} & $s{=}20$ & $s{=}40$ & $s{=}60$ & $s{=}100$ \\
\midrule
\multicolumn{6}{c}{\textit{Log-Probability Drop} $\uparrow$} \\
\midrule
\multirow{3}{*}{CAMAB}
    & 1 & 0.501 & 0.521 & \textbf{0.526} & \textbf{0.526} \\
    & 3 & 0.617 & 0.676 & 0.691 & \textbf{0.693} \\
    & 5 & 0.647 & 0.717 & \textbf{0.733} & 0.730 \\
\midrule
\multirow{3}{*}{SHAP}
    & 1 & 0.265 & 0.475 & 0.509 & 0.520 \\
    & 3 & 0.362 & 0.614 & 0.666 & 0.704 \\
    & 5 & 0.414 & 0.648 & 0.710 & 0.747 \\
\midrule
\multicolumn{6}{c}{\textit{BERTScore} $\downarrow$} \\
\midrule
\multirow{3}{*}{CAMAB}
    & 1 & 0.525 & \textbf{0.509} & 0.511 & 0.521 \\
    & 3 & 0.464 & 0.421 & \textbf{0.418} & 0.429 \\
    & 5 & 0.445 & 0.407 & \textbf{0.402} & 0.404 \\
\midrule
\multirow{3}{*}{SHAP}
    & 1 & 0.668 & 0.562 & 0.527 & 0.529 \\
    & 3 & 0.598 & 0.471 & 0.444 & 0.440 \\
    & 5 & 0.562 & 0.453 & 0.423 & 0.422 \\
\bottomrule
\end{tabular}
\end{table}

\subsection{Additional Results}
From Table~\ref{tab:eval_qwen}, for the Qwen-2.5-7B model, CAMAB consistently outperforms both SHAP and ContextCite across nearly all datasets and metrics.

\subsection{Interaction Effects in the Precision Matrix}
\label{sec:interaction}

A potential concern with the linear reward assumption (Eq.~\ref{eq:reward_linear}) is that it may miss synergies between context segments. Here we demonstrate empirically that the posterior precision matrix $\boldsymbol{B}_t$ captures such interactions implicitly.

\begin{figure*}[t]
    \centering
    \includegraphics[width=\textwidth]{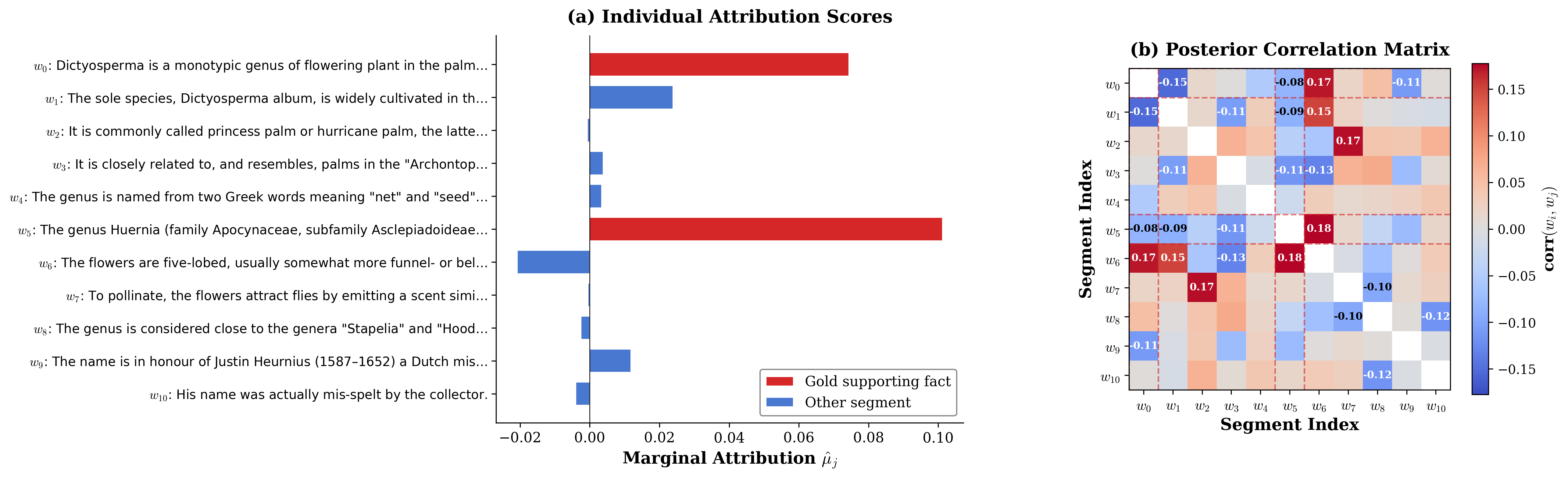}
    \caption{Posterior analysis on a HotpotQA example (``Are both Dictyosperma and Huernia described as a genus?''). \textbf{Left}: marginal attribution $\hat{\mu}_j$ per segment; gold supporting facts (segments 0 and 5) are the top-ranked. \textbf{Right}: posterior correlation matrix (diagonal masked); the strongest off-diagonal entries cluster within each sub-topic, revealing interaction effects that marginal scores alone cannot capture.}
    \label{fig:heatmap}
\end{figure*}

Figure~\ref{fig:heatmap} visualizes the posterior on a HotpotQA example with 11 segments and 2 gold supporting facts, after $n_\text{iter}{=}200$ rounds of LinTS. The left panel shows the marginal attribution $\hat{\mu}_j$ per segment; the two gold supporting facts (segments 0 and 5) are correctly identified as the top-ranked. The right panel displays the posterior correlation matrix (with the diagonal masked). The strongest off-diagonal entries cluster \textit{within} each sub-topic---for instance, corr($w_5, w_6$) = +0.18 links two Huernia-related segments---capturing substitutability effects that marginal attribution scores alone cannot express.

\end{document}